\documentclass[10pt,twocolumn]{article}

\usepackage[margin={1.8cm,1.8cm}]{geometry}

\usepackage{times}

\usepackage{placeins}
\usepackage{epsfig}
\usepackage{graphicx}
\usepackage{amsmath}
\usepackage{amssymb}
\usepackage{subfigure}
\usepackage{booktabs}
\usepackage{multirow}
\usepackage{xspace}
\usepackage{array}
\usepackage[usenames]{color}
\usepackage{bm}
\usepackage{ragged2e}
\usepackage{icomma}
\usepackage{hyphenat}
\usepackage{adjustbox}

\usepackage[breaklinks=true,letterpaper=true,colorlinks,bookmarks=false]{hyperref}

\newcommand{\etal}{{et al.~}}
\newcommand{\ie}{{i.e.,~}}

\definecolor{shadecolour}{gray}{0.4}

\newcommand{\set}[1]{{\mathcal #1}}
\newcommand{\mat}[1]{\boldsymbol{\mathbf{#1}}} 
\renewcommand{\vec}[1]{\boldsymbol{\mathbf{#1}}} 

\newcommand{\D}{{\mat{D}}}
\newcommand{\U}{{\mat{Y}}}
\newcommand{\M}{{\mat{M}}}
\newcommand{\E}{{\set{R}}}
\newcommand{\NegSet}{{\set{Q}}}
\newcommand{\C}{{\set{C}}}
\newcommand{\X}{{\mat{X}}}
\newcommand{\train}{{\set{U}}}

\newcommand{\x}{{\vec{x}}}
\newcommand{\s}{{\vec{s}}}
\newcommand{\w}{{\vec{w}}}
\newcommand{\rij}{r_{ij}}
\newcommand{\rqj}{r_{qj}}
\renewcommand{\P}{P}

\newcommand{\xhdr}[1]{\vspace{0mm}\noindent{{\bf #1}}}
\newcommand{\eq}[1]{(eq.~\ref{#1})}

\DeclareMathOperator*{\argmax}{\mathrm{argmax}}

\newcommand{\Amazon}{Amazon\xspace}

\newfont{\mycrnotice}{ptmr8t at 7pt}
\newfont{\myconfname}{ptmri8t at 7pt}

\usepackage{authblk}

\begin{document}

\title{Image-based Recommendations on Styles and Substitutes}

\author[1]{Julian McAuley\thanks{jmcauley@ucsd.edu}}
\author[2]{Christopher Targett\thanks{christopher.targett@student.adelaide.edu.au}}
\author[2]{Qinfeng (`Javen') Shi\thanks{javen.shi@adelaide.edu.au}}
\author[2,3]{Anton van den Hengel\thanks{anton.vandenhengel@adelaide.edu.au}}
\affil[1]{Department of Computer Science, UC San Diego}
\affil[2]{School of Computer Science, University of Adelaide}
\affil[3]{Australian Centre for Robot Vision}

\maketitle

\begin{abstract}
Humans inevitably develop a sense of the relationships between objects, some of which are based on their appearance.  Some pairs of objects might be seen as being alternatives to each other (such as two pairs of jeans), while others may be seen as being complementary (such as a pair of jeans and a matching shirt). This information guides many of the choices that people make, from buying clothes to their interactions with each other.  We seek here to model this human sense of the relationships between objects based on their appearance. Our approach is not based on fine-grained modeling of user annotations but rather on capturing the largest dataset possible and developing a scalable method for uncovering human notions of the visual relationships within. We cast this as a network inference problem defined on graphs of related images, and provide a large-scale dataset for the training and evaluation of the same.  The system we develop is capable of recommending which clothes and accessories will go well together (and which will not), amongst a host of other applications.
\end{abstract}

\section{Introduction}
\label{sec:intro}

We are interested here in uncovering relationships between the appearances of pairs of objects, and particularly in modeling the human notion of which objects complement each other and which might be seen as acceptable alternatives.  We thus seek to model what is a fundamentally human notion of the visual relationship between a pair of objects, rather than merely modeling the visual similarity between them.  There has been some interest of late in modeling the visual style of places \cite{doersch2012what,Knowing2}, and objects \cite{Xu:2010:SSA:1882261.1866206}.  We, in contrast, are not seeking to model the individual appearances of objects, but rather how the appearance of one object might influence the desirable visual attributes of another.

\begin{figure}[t]
	\centering
		\includegraphics[height=1.6cm]{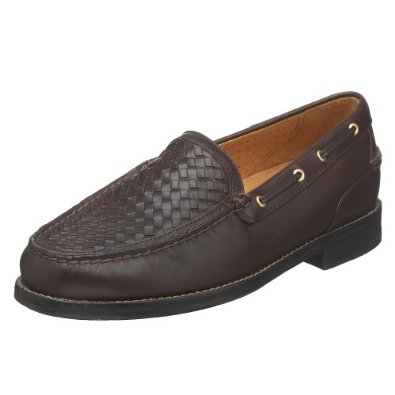}\hspace{1mm}\vline\hspace{1mm}
		\includegraphics[height=1.6cm]{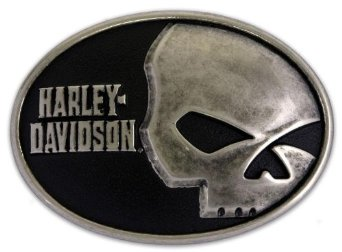}
		\includegraphics[height=1.6cm]{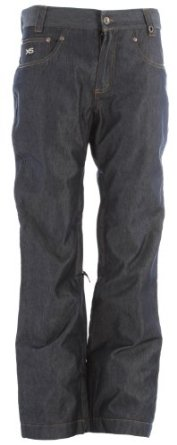}
		\includegraphics[height=1.6cm]{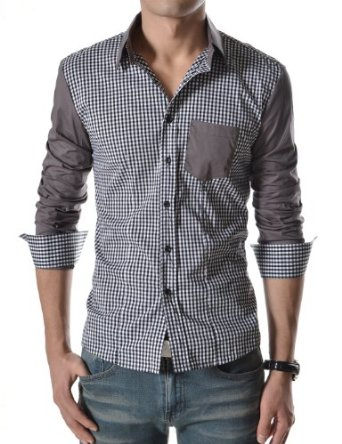}
	\caption{A query image and a matching accessory, pants, and a shirt.}
	\label{fig:brownLoaferQuery}
\end{figure}

There are a range of situations in which the appearance of an object might have an impact on the desired appearance of another.  
Questions such as `Which frame goes with this picture', `Where is the lid to this', and  `Which shirt matches these shoes' (see Figure~\ref{fig:brownLoaferQuery}) inherently involve a calculation of more than just visual similarity, but rather a model of the higher-level relationships between objects.
The primary commercial application for such technology is in recommending items to a user based on other items they have already showed interest in.
Such systems are of considerable economic value, and are typically built by analysing meta-data, reviews, and previous purchasing patterns.  By introducing into these systems the ability to examine the appearance of the objects in question we aim to overcome some of their limitations, including the `cold start' problem \cite{schein,ke}.

The problem we pose inherently requires modeling human visual preferences.  In most cases there is no intrinsic connection between a pair of objects, only a human notion that they are more suited to each other than are other potential partners.  The most common approach to modeling such human notions exploits a set of hand-labeled images created for the task.  The labeling effort required means that most such datasets are typically relatively small, although there are a few notable exceptions.  A small dataset means that complex procedures are required to extract as much information as possible without over-fitting (see \cite{carneiro2007supervised,Di:2013,makadia2008new} for example).  It also means that the results are unlikely to be transferable to related problems.
Creating a labeled dataset is particularly onerous when modeling pairwise distances because the number of annotations required scales with the square of the number of elements.

We propose here instead that one might operate over a much larger dataset, even if it is only tangentially related to the ultimate goal.  Thus, rather than devising a process (or budget) for manually annotating images, we instead seek a freely available source of a large amount of data which may be more loosely related to the information we seek.  Large-scale databases have been collected from the web (without other annotation) previously \cite{hays2008im2gps,torralba200880}. 
What distinguishes the approach we propose here, however, is the fact that it succeeds despite the indirectness of the connection between the dataset and the quantity we hope to model. 

\subsection{A visual dataset of styles and substitutes}
\label{sec:data}

We have developed a dataset suitable for the purposes described above based on the \Amazon web store.
The dataset contains over $180$ million 
relationships between a pool of almost 6~million objects.  These relationships are a result of visiting \Amazon and recording the product recommendations that it provides given our (apparent) interest in the subject of a particular web page.  
The statistics of the dataset are shown in Table \ref{tab:stats}.
 An image and a category label are available for each object, as is the set of users who reviewed it.
We have made this dataset available for academic use, along with all code used in this paper to ensure that our results are reproducible and extensible.\footnote{\url{http://cseweb.ucsd.edu/~jmcauley/}}
We label this the \emph{Styles and Substitutes} dataset.

The recorded relationships describe
two specific notions of `compatibility' that are of interest, namely those of \emph{substitute} and \emph{complement} goods. Substitute goods are those that can be interchanged (such as one pair of pants for another), while complements are those that might be purchased together (such as a pair of pants and a matching shirt) \cite{micro}.
Specifically, there are 4 categories of relationship represented in the dataset: 1) `users who viewed X also viewed Y' (65M edges); 2) `users who viewed X eventually bought Y' (7.3M edges); 3) `users who bought X also bought Y' (104M edges); and 4) `users bought X and Y simultaneously' (3.4M edges). Critically, categories 1 and 2 indicate (up to some noise) that two products may be substitutable, while 3 and 4 indicate that two products may be complementary.
According to \Amazon's own tech report \cite{Linden03} the above relationships are collected simply by ranking products according to the cosine similarity of the sets of users who purchased/viewed them.

Note that the dataset does not document users' preferences for pairs of images, but rather \Amazon's estimate of the set of relationships between pairs objects.  
The human notion of the visual compatibility of these images is only one factor amongst many which give rise to these estimated relationships, and it is not a factor used by \Amazon in creating them.  
We thus do not wish to summarize the \Amazon data, but rather to use what it tells us about the images of related products to develop a sense of which objects a human might feel are visually compatible.  This is significant because many of the relationships between objects present in the data are not based on their appearance.  People co-purchase hammers and nails due to their functions, for example, not their appearances.
Our hope is that the non-visual decision factors will appear as uniformly distributed noise to a method which considers only appearance, and that the visual decision factors might reinforce each other to overcome the effect of this noise.

\begin{table*}
\begin{center}
\small
 \begin{tabular}{lrrrr}
\toprule
Category & Users & Items & Ratings & Edges\\
\midrule
Books 				& 8,201,127 & 1,606,219 & 25,875,237 & 51,276,522\\
Cell Phones \& Accessories	& 2,296,534 & 223,680	& 5,929,668 & 4,485,570\\
Clothing, Shoes \& Jewelry	& 3,260,278 & 773,465	& 25,361,968 & 16,508,162\\
Digital Music	 		& 490,058 & 91,236	& 950,621 & 1,615,473\\
Electronics	 		& 4,248,431 & 305,029	& 11,355,142 & 7,500,100\\
Grocery \& Gourmet Food		& 774,095 & 120,774	& 1,997,599 & 4,452,989\\
Home \& Kitchen	 		& 2,541,693 & 282,779	& 6,543,736 & 9,240,125\\
Movies \& TV	 		& 2,114,748 & 150,334	& 6,174,098 & 5,474,976\\
Musical Instruments		& 353,983 & 65,588	& 596,095 & 1,719,204\\
Office Products	 		& 919,512 & 94,820	& 1,514,235 & 3,257,651\\
Toys \& Games	 		& 1,352,110 & 259,290	& 2,386,102 & 13,921,925\\
\midrule
Total				& 20,980,320 & 5,933,184	& 143,663,229 & 180,827,502\\
\bottomrule
\end{tabular}
\normalsize
\end{center}
 \caption{The types of objects from a few categories in our dataset and the number of relationships between them. \label{tab:stats}}
\end{table*}

\subsection{Related work}

The closest systems to what we propose above are content-based recommender systems \cite{lew2006content} which attempt to model each user's preference toward particular types of goods.  This is typically achieved by analyzing metadata from the user's previous activities.  This is as compared to collaborative recommendation approaches which match the user to profiles generated based on the purchases/behavior of other users (see \cite{adomavicius2005toward,korenSurvey} for surveys).  
Combinations of the two \cite{bilinear,melville2002content} have been shown to help address the sparsity of the review data available, and the cold-start problem (where new products don't have reviews and are thus invisible to the recommender system) \cite{schein,ke}.  The approach we propose here could also help address these problems.

There are a range of services such as Jinni\footnote{\url{http://jinni.com}} which promise content-based recommendations for TV shows and similar media, but the features they expoit are based on reviews and meta-data (such as cast, director etc.), and their ontology is hand-crafted.  The Netflix prize was a well publicized competition to build a better personalized video recommender system, but there again no actual image analysis is taking place \cite{koren2009matrix}. 
Hu \etal \cite{Etsy} describe a system for identifying a user's style, and then making clothing recommendations, but this is achieved through analysis of `likes' rather than visual features.

Content-based image retrieval gives rise to the problem of bridging the `semantic-gap' \cite{smeulders2000content}, which requires returning results which have similar semantic content to a search image, even when the pixels bear no relationship to each other.  It thus bears some similarity to the visual recommendation problem, as both require modeling a human preference which is not satisfied by mere visual similarity.
There are a variety of approaches to this problem, many of which seek a set of results which are visually similar to the query and then separately find images depicting objects of the same class as those in the query image; see~\cite{carneiro2007supervised,hashing1,makadia2008new,kernel1}, for example. Within the Information Retrieval community there has been considerable interest of late in incorporating user data into image retrieval systems \cite{learning2}, for example  through browsing \cite{ranking1} and click-through behavior \cite{clickthrough}, or by making use of social tags~\cite{prism}.
Also worth mentioning with respect to image retrieval is \cite{amazonJaiwei}, which also considered using images crawled from Amazon, albeit for a different task (similar-image search) than the one considered here.

There have been a variety of approaches to modeling human notions of similarity between different types of images \cite{shrivastava-sa11}, forms of music \cite{slaney2008learning}, or even tweets \cite{learning1}, amongst other data types.
Beyond measuring similarity, there has also been work on measuring more general notions of compatibility. Murillo \etal \cite{murillo2012urban}, for instance,  analyze photos of groups of people collected from social media to identify which groups might be more likely to socialize with each other, thus implying a distance measure between images.  This is achieved by estimating which of a manually-specified set of `urban tribes' each group belongs to, possibly because only 340 images were available. 

Yamaguchi \etal \cite{yamaguchi2013paper} capture a notion of visual style when parsing clothing, but do so by retrieving visually similar items from a database.  This idea was extended by Kiapour \etal \cite{kiapour2014hipster} to identify discriminating characteristics between different styles (hipster vs.~goth for example). Di \etal \cite{Di:2013} also identify aspects of style using a bag-of-words approach and manual annotations.

A few other works that consider visual features specifically for the task of clothing recommendation include~\cite{streetFashion,gettingTheLook,magicCloset}. In \cite{streetFashion} and \cite{gettingTheLook} the authors build methods to parse complete outfits from single images, in \cite{streetFashion} by building a carefully labeled dataset of street images annotated by `fashionistas', and in \cite{gettingTheLook} by building algorithms to automatically detect and segment items from clothing images. In \cite{gettingTheLook} the authors propose an approach to learn relationships between clothing items and events (e.g.~birthday parties, funerals) in order to recommend event-appropriate items. Although related to our approach, these methods are designed for the specific task of clothing recommendation, requiring hand-crafted methods and carefully annotated data; in contrast our goal is to build a general-purpose method to understand relationships between objects from large volumes of \emph{unlabeled} data. Although our setting is perhaps most natural for categories like clothing images, we obtain surprisingly accurate performance when predicting relationships in a variety of categories, from recommending outfits to predicting which books will be co-purchased based on their cover art.

In summary, our approach is distinct from the above in that we aim to generalize the idea of a visual distance measure beyond measuring only similarity.  Doing so demands a very large amount of training data, and our reluctance for manual annotation necessitates a more opportunistic data collection strategy.  The scale of the data, and the fact that we don't have control over its acquisition, demands a suitably scalable and robust modeling approach.  The novelty in what we propose is thus in the quantity we choose to model, the data we gather to do so, and the method for extracting one from the other.

\subsection{A visual and relational recommender system}
\label{sec:visual}

We label the process we develop for exploiting this data a \emph{visual and relational recommender system} as we aim to model human visual preferences, and 
the system might be used to recommend one object on the basis of a user's apparent interest in another.  
The system shares these characteristics with more common forms of recommender system, but does so on the basis of the appearance of the object, rather than metadata, reviews, or similar. 

\section{The Model}
\label{sec:Model}

\begin{table}
\begin{center}
\small
\begin{tabular}{m{0.14\linewidth}m{0.74\linewidth}}
 \toprule
 \bf notation & \bf explanation\\
 \midrule
 $\x_i$ & feature vector calculated from object image $i$ \\
 $F$ & feature dimension (\ie $\x_i \in {\mathbb R}^F$)\\
 $\rij$ & a relationship between objects $i$ and $j$\\
 $\E$ & the set of relationships between all objects\\
 $d_\theta(\x_i,\x_j)\!\!\!\!$ & parameterized distance between $\x_i$ and $\x_j$\\
  $\M$ & $F \times F$ Mahalanobis transform matrix\\
 $\U$ & an $F \times K$ matrix, such that $\U\U^T = \M$\\
 $\D^{(u)}$ & diagonal user-personalization matrix for user $u$\\
 $\sigma_c(\cdot)$ & shifted sigmoid function with parameter $c$\\
 $\E^*$ & $\E$ plus a random sample of non-relationships\\
 $\train, \mathcal V, \mathcal T$ & training, validation, and test subsets of $\E^*$\\
 $\s_i$ & $K$-dimension embedding of $\x_i$ into `style-space'\\
 \bottomrule
\end{tabular}
\normalsize
\end{center}
 \caption{Notation.\label{tab:notation}}
\end{table}

Our notation is defined in Table \ref{tab:notation}.

We seek a method for representing the preferences of users for the visual appearance of one object given that of another.  A number of suitable models might be devised for this purpose, but very few of them will scale to the volume of data available.  

For every object in the dataset we calculate an $F$-dimensio\-nal feature vector $\x \in {\mathbb R}^F$ using a convolutional neural network as described in Section~\ref{sec:features}.
The dataset contains a set $\E$ of relationships where $\rij \in \E$ relates objects $i$ and $j$. Each relationship is of one of the four classes listed above.
Our goal is to learn a parameterized distance transform $d(\x_i,\x_j)$ such that feature vectors $\{\x_i,\x_j\}$ for objects that are related ($\rij \in \E$) are assigned a lower distance than those that are not ($\rij \notin \E$). Specifically, we seek $d(\cdot,\cdot)$ such that
 $\P(\rij \in \E)$ grows monotonically with $-d(\x_i,\x_j)$.

\begin{figure}[t]
 \includegraphics[width=0.8\linewidth]{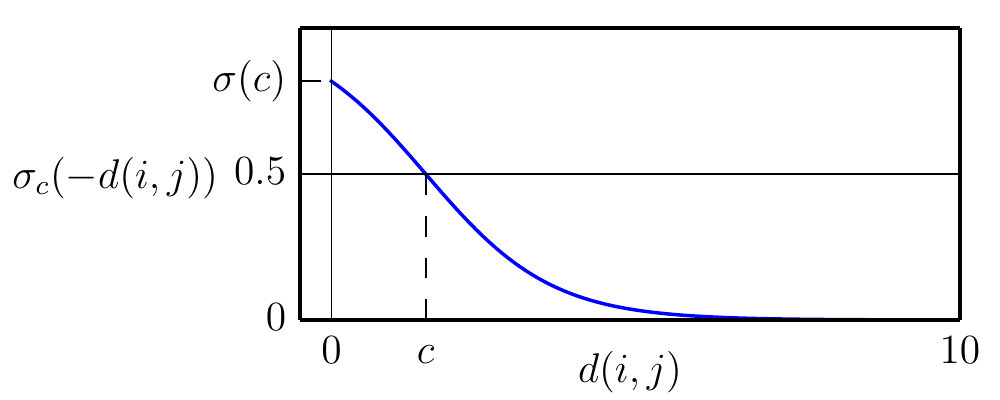}
\caption{Shifted (and inverted) sigmoid with parameter $c=2$.\label{fig:sigmoid}}
\end{figure}

\xhdr{Distances and probabilities:}
We use a shifted sigmoid function to relate distance to probability thus
\begin{equation}
 \P(\rij \in \E) = \sigma_c(-d(\x_i,\x_j)) = \frac{1}{1 + e^{d(\x_i,\x_j) - c}}.
 \label{eq:prob}
\end{equation}
This is depicted in Figure \ref{fig:sigmoid}. 
This decision allows us to cast the problem as logistic regression, which we do for reasons of scalability.
Intuitively, if two items $i$ and $j$ have distance $d(\x_i,\x_j) = c$, then they have probability $0.5$ of being related; the probability increases above $0.5$ for $d(\x_i,\x_j) < c$, and decreases as $d(\x_i,\x_j) > c$. 
Note that we do not specify $c$ in advance, but rather $c$ is chosen to maximize prediction accuracy.

We now describe a set of potential distance functions.

\xhdr{Weighted nearest neighbor:} 
Given that different feature dimensions are likely to be more important to different relationships, the simplest method we consider is to learn which feature dimensions are relevant for a particular relationship. We thus fit a distance function of the form
\begin{equation}
 d_\w(\x_i,\x_j) =\ \| \w \circ (\x_i-\x_j) \|_2^2,
 \label{eq:theta}
\end{equation}
where $\circ$ is the Hadamard product.

\xhdr{Mahalanobis transform:} \eq{eq:theta} is limited to modeling the visual similarity between objects, albeit with varying emphasis per feature dimension. It is not expressive enough to model subtler notions, such as which pairs of pants and shoes belong to the same `style', despite having different appearances. For this we need to learn how different feature dimensions \emph{relate} to each other, i.e., how the features of a pair of pants might be transformed to help identify a compatible pair of shoes.

To identify such a transformation, we relate image features via a \emph{Mahalanobis distance}, which essentially generalizes \eq{eq:theta} so that weights are defined at the level of pairs of features. Specifically we fit
\begin{equation}
    d_\M(\x_i,\x_j)  = (\x_i - \x_j)\M(\x_i - \x_j)^T.
  \label{eq:M}
\end{equation}

A full rank p.s.d.~matrix $\M$ has too many parameters to fit tractably given the size of the dataset. For example, using features with dimension $F = 2^{12}$, learning a transform as in \eq{eq:M} requires us to fit 
approximately $8$~million parameters; not only would this be prone to overfitting,  it is simply not practical for existing solvers.

To address these issues, and given the fact that $\M$ parameterises a Mahanalobis distance, we approximate $\M$ such that $\M \simeq \U\U^T$ where $\U$ is a matrix of dimension $F \times K$. We therefore define
\begin{equation}
 \begin{aligned}
  d_\U(\x_i,\x_j) & = (\x_i - \x_j)\U\U^T(\x_i - \x_j)^T\\
           & = \| (\x_i - \x_j)\U \|_2^2.
 \end{aligned}
 \label{eq:U}
\end{equation}
Note that
all distances (as well as their derivatives) can be computed in $O(FK)$, which is significant for the scalability of the method. Similar ideas appear in \cite{Der12,Tor07}, which also consider the problem of metric learning via low-rank embeddings, albeit using a different objective than the one we consider here.

\subsection{Style space}

In addition to being computationally useful, the low-rank transform in \eq{eq:U} has a convenient interpretation. Specifically, if we consider the $K$-dimensional vector
$
 \s_i = \x_i\U
$,
then \eq{eq:U} can be rewritten as
\begin{equation}
 d_\U(\x_i,\x_j) = \| \s_i - \s_j \|^2_2.
 \label{eq:styledist}
\end{equation}
In other words, \eq{eq:U} yields a low-dimensional embedding of the features $\x_i$ and $\x_j$. We refer to this low-dimensional representation as the product's embedding into `style-space', in the hope that we might identify $\U$ such that related objects fall close to each other despite being visually dissimilar. 
The notion of `style' is learned automatically by training the model on pairs of objects which \Amazon considers to be related.  

\subsection{Personalizing styles to individual users}

So far we have developed a model to learn a \emph{global} notion of which products go together, by learning a notion of `style' such that related products should have similar styles. As an addition to this model we can personalize this notion by learning for each individual user which dimensions of style they consider to be important.

To do so, we shall learn personalized distance functions $d_{\U,u}(\mathbf{x}_i, \mathbf{x}_j)$ that measure the distance between the items $i$ and $j$ \emph{according to the user $u$}. We choose the distance function
\begin{equation}
 d_{\U,u}(\mathbf{x}_i, \mathbf{x}_j) = (\mathbf{x}_i - \mathbf{x}_j)\U\D^{(u)}\U^T(\mathbf{x}_i - \mathbf{x}_j)^T
 \label{eq:personal1}
\end{equation}
where $\D^{(u)}$ is a $K\times K$ diagonal (positive semidefinite) matrix. In this way the entry $\D^{(u)}_{kk}$ indicates the extent to which the user $u$ `cares about' the $k^{\text{th}}$ style dimension.

In practice we fit a $U\times K$ matrix $\X$ such that $D^{(u)}_{kk} = \X_{uk}$. Much like the simplification in \eq{eq:styledist}, the distance $d_{\U,u}(\mathbf{x}_i, \mathbf{x}_j)$ can be conveniently written as
\begin{equation}
 d_{\U,u}(\mathbf{x}_i, \mathbf{x}_j) = \| (\mathbf{s}_i - \mathbf{s}_j) \circ X_{u} \|^2_2.
 \label{eq:personal2}
\end{equation}
In other words, $X_u$ is a personalized \emph{weighting} of the projected style-space dimensions.

The construction in (eq.~\ref{eq:personal1} and \ref{eq:personal2}) only makes sense if there are \emph{users} associated with each edge in our dataset, which is not true of the four graph types we have presented so far. Thus to study the issue of user personalization we make use of our rating and review data (see Table \ref{tab:stats}). From this we sample a dataset of triples $(i,j,u)$ of products $i$ and $j$ that were both purchased by user $u$ (i.e., $u$ reviewed them both). We describe this further when we outline our experimental protocol in Section \ref{sec:protocol}.

\subsection{Features}
\label{sec:features}

Features are calculated from the original images using the Caffe deep learning framework \cite{jia2014caffe}.  In particular, we used a Caffe reference model\footnote{bvlc\_reference\_caffenet from \url{caffe.berkeleyvision.org}} with 5 convolutional layers followed by 3 fully-connected layers, which has been pre-trained on $1.2$ million ImageNet (ILSVRC2010) images.  We use the output of FC7, the second fully-connected layer, which results in a feature vector of length $F=4096$. 

\section{Training}
\label{sec:training}

Since we have defined a probability associated with the presence (or absence) of each relationship, we can proceed by maximizing the likelihood of an observed relationship set $\E$. 
In order to do so we randomly select a negative set $\NegSet = \{\rij | \rij \notin \E\}$ such that $|\NegSet| = |\E|$ and optimize the log likelihood
\begin{multline}
 l(\U,c | \E, \NegSet) = \sum_{\rij\in \E} \log(\sigma_c(-d_\U(\x_i,\x_j))) +\\ \sum_{\rij \in \NegSet} \log(1 - \sigma_c(-d_\U(\x_i,\x_j))).
\label{eq:likelihood}
\end{multline}
Learning then proceeds by optimizing $l(\U,c | \E, \NegSet)$ over both $\U$ and $c$
which we achieve by gradient ascent. We use (hybrid) L-BFGS, a quasi-Newton method for non-linear optimization of problems with many variables \cite{hlbfgs}. Likelihood \eq{eq:likelihood} and derivative computations can be na\"ively parallelized over all pairs $\rij\in \E\cup\NegSet$. Training on our largest dataset (\Amazon books) with a rank $K=100$ transform required around one day on a 12 core machine.

\section{Experiments}
\label{sec:Ex}

We compare our model against the following baselines:

We compare against \emph{Weighted Nearest Neighbor (WNN)} classification, as is described in Section \ref{sec:visual}.
We also compare against a method we label \emph{Category Tree (CT)}; CT is based on using
\Amazon's detailed category tree directly (which we have collected for Clothing data, and use for later experiments), which allows us to assess how effective an image-based classification approach \emph{could} be, if it were perfect. We then compute a matrix of 
coocurrences
between categories from the training data, and label two products $(a,b)$ as `related' if the category of $b$ belongs to one of the top 50\% of most commonly linked categories for products of category $a$.\footnote{We experimented with several variations on this theme, and this approach yielded the best performance.}
Nearest neighbor results (calculated by optimizing a threshold on the $\ell_2$ distance using the training data) were not significantly better than random, and have been suppressed for brevity.

\xhdr{Comparison against non-visual baselines} As a non-visual comparison, we trained topic models on the reviews of each product (i.e., each document $d_i$ is the set of reviews of the product $i$) and fit weighted nearest neighbor classifiers of the form
\begin{equation}
 d_\w(\theta_i, \theta_j) = \|\w \circ (\theta_i - \theta_j)\|_2^2,
\end{equation}
where $\theta_i$ and $\theta_j$ are topic vectors derived from the reviews of the products $i$ and $j$. In other words, we simply adapted our WNN baseline to make use of topic vectors rather than image features.\footnote{We tried the same approach at the word (rather than the topic) level, though this led to slightly worse results.} We used a 100-dimensional topic model trained using Vowpal Wabbit \cite{Hoffman10onlinelearning}.

However, this baseline proved not to be competitive against the alternatives described above (e.g.~only 60\% accuracy on our largest dataset, `Books'). One explanation may simply be that is is difficult to effectively train topic models at the 1M+ document scale; another explanation is simply that the vast majority of products have few reviews. Not surprisingly, the number of reviews per product follows a power-law, e.g.~for Men's Clothing:
\begin{center}
 \includegraphics[width=2in]{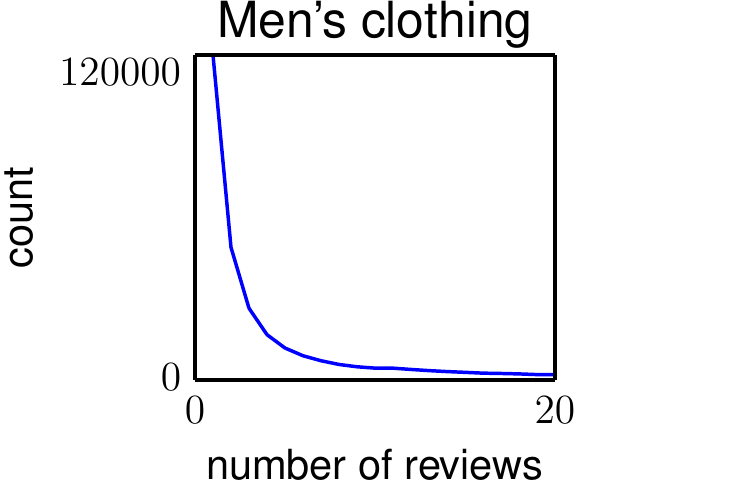}
\end{center}
This issue is in fact exacerbated in our setting, as to predict a relationship between products we require both to have reliable feature representations, which will be true only if \emph{both} products have several reviews.

Although we believe that predicting such relationships using text is a promising direction of future research (and one we are exploring), we simply wish to highlight the fact that there appears to be no `silver bullet' to predict such relationships using text, primarily due to the `cold start' issue that arises due to the long tail of obscure products with little text associated with them. Indeed, this is a strong argument in favor of building predictors based on visual features, since images are available even for brand new products which are yet to receive even a single review.

\subsection{Experimental protocol}
\label{sec:protocol}

We split the dataset into its top-level categories (Books, Movies, Music, etc.) and further split the Clothing category into second-level categories (Men's, Women's, Boys, Girls, etc.). We focus on results from a few representative subcategories. Complete code for all experiments and all baselines is available online.\footnote{\url{http://cseweb.ucsd.edu/~jmcauley/}}

\begin{table}[t]
 \begin{center}
  \begin{tabular}{llc}
  \toprule
  Category & method & accuracy\\
  \midrule
     \multirow{4}{21mm}{Men's clothing} & CT & 84.8\%\\
                                        & WNN & 84.3\%\\
                                        & $K=10$, no personalization & 90.9\%\\
                                        & $K=10$, personalized & 93.2\%\\[2mm]
     \multirow{4}{21mm}{Women's clothing} & CT & 80.5\%\\
                                          & WNN & 80.8\%\\
                                          & $K=10$, no personalization & 87.6\%\\
                                          & $K=10$, personalized & 89.1\%\\
                                          \bottomrule
  \end{tabular}
 \end{center}
 \caption{Performance of our model at predicting copurchases with a user personalization term (eqs.~\ref{eq:personal1} and \ref{eq:personal2}). \label{tab:personal}}
\end{table}

\begin{table}[t]
\renewcommand{\tabcolsep}{2pt}
\begin{center}
\small
\begin{tabular}{lrcccc}
\toprule
&&\multicolumn{2}{c}{\bf substitutes}&\multicolumn{2}{c}{\bf complements}\\
Category & method &\ \hspace{-2mm}\parbox{13.5mm}{\centering b\vphantom{hj}uy after viewin\vphantom{hj}g}\hspace{-2mm}\ & \ \hspace{-2mm}\parbox{13.5mm}{\centering a\vphantom{hj}lso viewe\vphantom{hj}d}\hspace{-2mm}\ &\ \hspace{-2mm}\parbox{13.5mm}{\centering a\vphantom{hj}lso bough\vphantom{hj}t}\hspace{-2mm}\ &\ \hspace{-2mm}\parbox{13.5mm}{\centering b\vphantom{hj}ought togethe\vphantom{hj}r}\hspace{-2mm}\ \\
\midrule
\multirow{3}{21mm}{Books}
 & WNN & 66.5\% & 62.8\% & 63.3\% & 65.4\% \\
 & $K = 10$ & 70.1\% & 68.6\% & 69.3\% & 68.1\% \\
 & $K = 100$ & 71.2\% & 69.8\% & 71.2\% & 68.6\% \\[2mm]
\multirow{3}{21mm}{Cell Phones and Accessories}
 & WNN & 73.4\% & 66.4\% & 69.1\% & 79.3\% \\
 & $K = 10$ & 84.3\% & 78.9\% & 78.7\% & 83.1\% \\
 & $K = 100$ & 85.9\% & 83.1\% & 83.2\% & 87.7\% \\[2mm]
\multirow{3}{21mm}{Clothing, Shoes,\\ and Jewelry}
 & WNN & $\cdot$ & 77.2\% & 74.2\% & 78.3\% \\
 & $K = 10$ & $\cdot$ & 87.5\% & 84.7\% & 89.7\% \\
 & $K = 100$ & $\cdot$ & 88.8\% & 88.7\% & 92.5\% \\[2mm]
\multirow{3}{21mm}{Digital Music}
 & WNN & 60.2\% & 56.7\% & 62.2\% & 53.3\% \\
 & $K = 10$ & 68.7\% & 60.9\% & 74.7\% & 56.0\% \\
 & $K = 100$ & 72.3\% & 63.8\% & 76.2\% & 59.0\% \\[2mm]
\multirow{3}{21mm}{Electronics}
 & WNN & 76.5\% & 73.8\% & 67.6\% & 73.5\% \\
 & $K = 10$ & 83.6\% & 80.3\% & 77.8\% & 79.6\% \\
 & $K = 100$ & 86.4\% & 84.0\% & 82.6\% & 83.2\% \\[2mm]
\multirow{3}{21mm}{Grocery and\\ Gourmet Food}
 & WNN & $\cdot$ & 69.2\% & 70.7\% & 68.5\% \\
 & $K = 10$ & $\cdot$ & 77.8\% & 81.2\% & 79.6\% \\
 & $K = 100$ & $\cdot$ & 82.5\% & 85.2\% & 84.5\% \\[2mm]
\multirow{3}{21mm}{Home and\\ Kitchen}
 & WNN & 75.1\% & 68.3\% & 70.4\% & 76.6\% \\
 & $K = 10$ & 78.5\% & 80.5\% & 78.8\% & 79.3\% \\
 & $K = 100$ & 81.6\% & 83.8\% & 83.4\% & 83.2\% \\[2mm]
\multirow{3}{21mm}{Movies and TV}
 & WNN & 66.8\% & 65.6\% & 61.6\% & 59.6\% \\
 & $K = 10$ & 71.9\% & 69.6\% & 72.8\% & 67.6\% \\
 & $K = 100$ & 72.3\% & 70.0\% & 77.3\% & 70.7\% \\[2mm]
\multirow{3}{21mm}{Musical\\ Instruments}
 & WNN & 79.0\% & 76.0\% & 75.0\% & 77.2\% \\
 & $K = 10$ & 84.7\% & 87.0\% & 85.3\% & 82.3\% \\
 & $K = 100$ & 89.5\% & 87.2\% & 84.4\% & 84.7\% \\[2mm]
\multirow{3}{21mm}{Office Products}
 & WNN & 72.8\% & 75.0\% & 74.4\% & 73.7\% \\
 & $K = 10$ & 81.2\% & 84.0\% & 84.1\% & 78.6\% \\
 & $K = 100$ & 85.9\% & 87.2\% & 85.8\% & 80.9\% \\[2mm]
\multirow{3}{21mm}{Toys and Games}
 & WNN & 67.0\% & 72.8\% & 71.7\% & 77.6\% \\
 & $K = 10$ & 75.8\% & 78.3\% & 78.4\% & 80.3\% \\
 & $K = 100$ & 77.1\% & 81.9\% & 82.4\% & 82.6\% \\
\bottomrule
\end{tabular}
\renewcommand{\tabcolsep}{6pt}
\normalsize
\end{center}
\caption{Accuracy of link prediction on top-level categories for each edge type with increasing model rank $K$. Random classification is 50\% accurate across all experiments. \label{tab:toplevel}}
\end{table}

\begin{table}[t]
\renewcommand{\tabcolsep}{4pt}
\begin{center}
\small
\begin{tabular}{lrccc}
\toprule
&&\multicolumn{1}{c}{\ \hspace{-3mm}\bf substitutes\hspace{-3mm}\ }&\multicolumn{2}{c}{\bf complements}\\
Category & method & \ \hspace{-2mm}\parbox{13.5mm}{\centering a\vphantom{hj}lso viewe\vphantom{hj}d}\hspace{-2mm}\ &\ \hspace{-2mm}\parbox{13.5mm}{\centering a\vphantom{hj}lso bough\vphantom{hj}t}\hspace{-2mm}\ &\ \hspace{-2mm}\parbox{13.5mm}{\centering b\vphantom{hj}ought togethe\vphantom{hj}r}\hspace{-2mm}\ \\
\midrule
\multirow{4}{21mm}{Baby}
 & CT & 77.1\% & 70.5\% & 80.1\% \\
 & WNN & 83.0\% & 87.7\% & 81.7\% \\
 & $K = 10$ & 92.2\% & 92.7\% & 91.5\% \\
 & $K = 100$ & 94.6\% & 94.3\% & 93.3\% \\[1.32mm]
\multirow{4}{21mm}{Boots}
 & CT & 75.0\% & 72.7\% & 74.2\% \\
 & WNN & 83.9\% & 85.6\% & 84.7\% \\
 & $K = 10$ & 93.0\% & 94.9\% & 95.4\% \\
 & $K = 100$ & 94.6\% & 96.8\% & 96.4\% \\[1.32mm]
\multirow{4}{21mm}{Boys}
 & CT & 81.9\% & 77.3\% & 83.1\% \\
 & WNN & 85.0\% & 87.2\% & 87.9\% \\
 & $K = 10$ & 94.4\% & 94.1\% & 93.8\% \\
 & $K = 100$ & 96.5\% & 95.8\% & 95.1\% \\[1.32mm]
\multirow{4}{21mm}{Girls}
 & CT & 83.0\% & 76.2\% & 78.7\% \\
 & WNN & 83.3\% & 86.0\% & 84.8\% \\
 & $K = 10$ & 94.5\% & 93.6\% & 93.0\% \\
 & $K = 100$ & 96.1\% & 95.3\% & 94.5\% \\[1.32mm]
\multirow{4}{21mm}{Jewelry}
 & CT & 50.1\% & 49.5\% & 51.1\% \\
 & WNN & 81.2\% & 81.6\% & 75.8\% \\
 & $K = 10$ & 89.6\% & 89.3\% & 82.8\% \\
 & $K = 100$ & 89.1\% & 91.6\% & 86.4\% \\[1.32mm]
\multirow{4}{21mm}{Men}
 & CT & 88.2\% & 78.4\% & 83.6\% \\
 & WNN & 86.9\% & 78.4\% & 82.3\% \\
 & $K = 10$ & 91.6\% & 89.8\% & 92.1\% \\
 & $K = 100$ & 92.6\% & 93.3\% & 95.1\% \\[1.32mm]
\multirow{4}{21mm}{Novelty\\ Costumes}
 & CT & 79.1\% & 76.3\% & 81.5\% \\
 & WNN & 80.1\% & 74.1\% & 76.0\% \\
 & $K = 10$ & 86.3\% & 86.6\% & 85.0\% \\
 & $K = 100$ & 89.2\% & 90.0\% & 89.1\% \\[1.32mm]
\multirow{4}{21mm}{Shoes and\\ Accessories}
 & CT & 81.3\% & 78.1\% & 90.4\% \\
 & WNN & 75.4\% & 80.2\% & 77.9\% \\
 & $K = 10$ & 89.7\% & 90.4\% & 93.5\% \\
 & $K = 100$ & 92.3\% & 94.7\% & 96.2\% \\[1.32mm]
\multirow{4}{21mm}{Women}
 & CT & 86.8\% & 79.1\% & 84.3\% \\
 & WNN & 78.8\% & 76.1\% & 80.0\% \\
 & $K = 10$ & 88.9\% & 87.8\% & 91.5\% \\
 & $K = 100$ & 90.4\% & 91.2\% & 94.3\% \\
\bottomrule
\end{tabular}
\renewcommand{\tabcolsep}{6pt}
\normalsize
\end{center}
\caption{Accuracy of link prediction on subcategories of `Clothing, Shoes, and Jewelry' with increasing rank $K$.  Note that `buy after viewing' links are not surfaced for clothing data on Amazon. \label{tab:clothing}}
\end{table}

\begin{figure}[t]
\renewcommand{\tabcolsep}{0mm}
\footnotesize
\vspace{-2mm}
\begin{center}
\includegraphics[width=\linewidth]{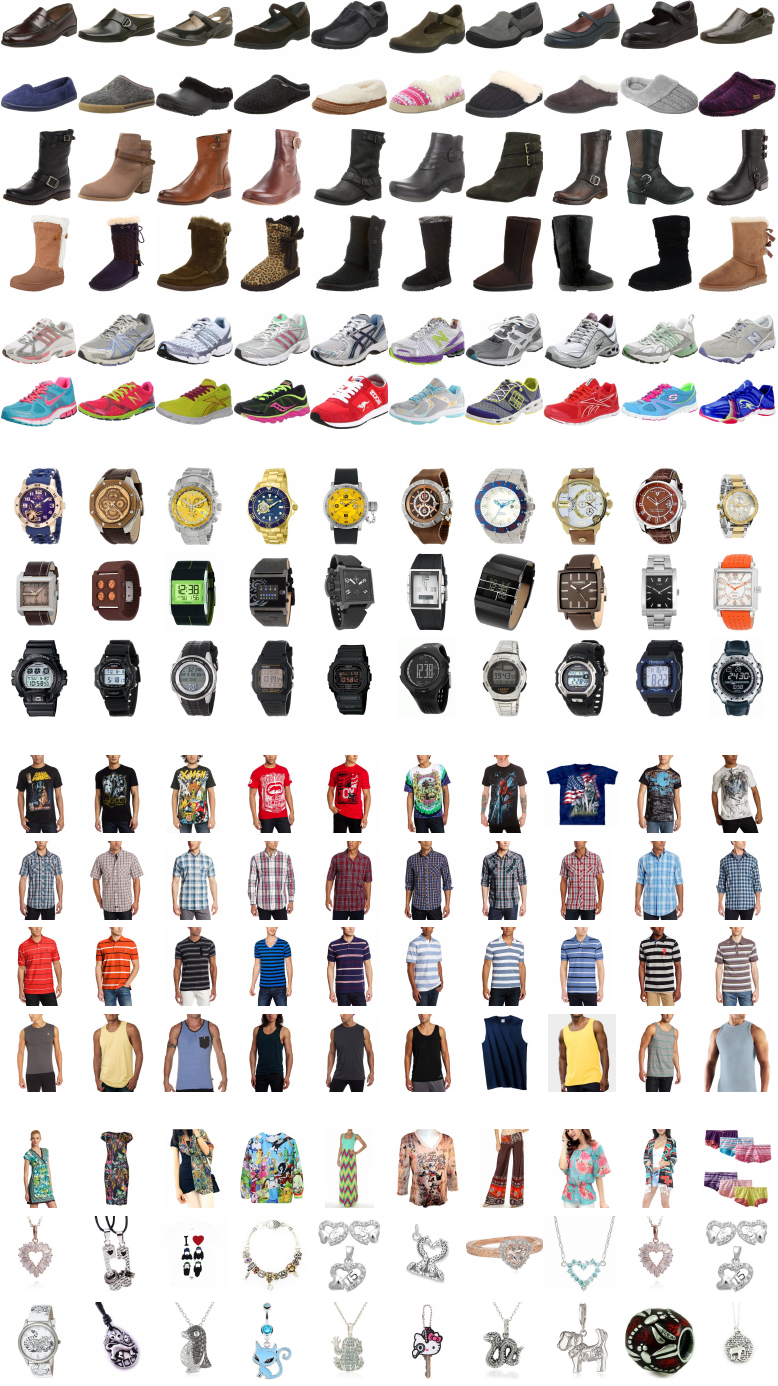}
\end{center}
\normalsize
\vspace{-5mm}
\caption{Examples of closely-clustered items in style space (Men's and Women's clothing `also viewed' data). \label{fig:pca}}
\end{figure}

\begin{figure}[t]
\setlength{\tabcolsep}{2pt}
	\centering
\includegraphics[width=\linewidth]{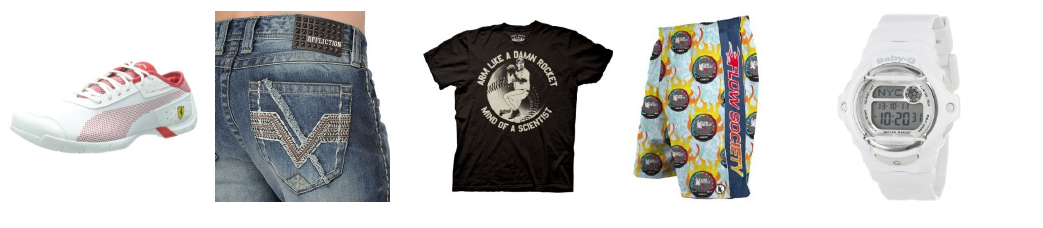}
	\vspace{-2mm}
	\caption{A selection of widely separated members of a single K-means cluster, demonstrating an apparent stylistic coherence.}
	\label{fig:WhiteShoe}
\end{figure}

\begin{figure}[t]
\renewcommand{\tabcolsep}{0mm}
\footnotesize
\begin{center}
\includegraphics[width=\linewidth]{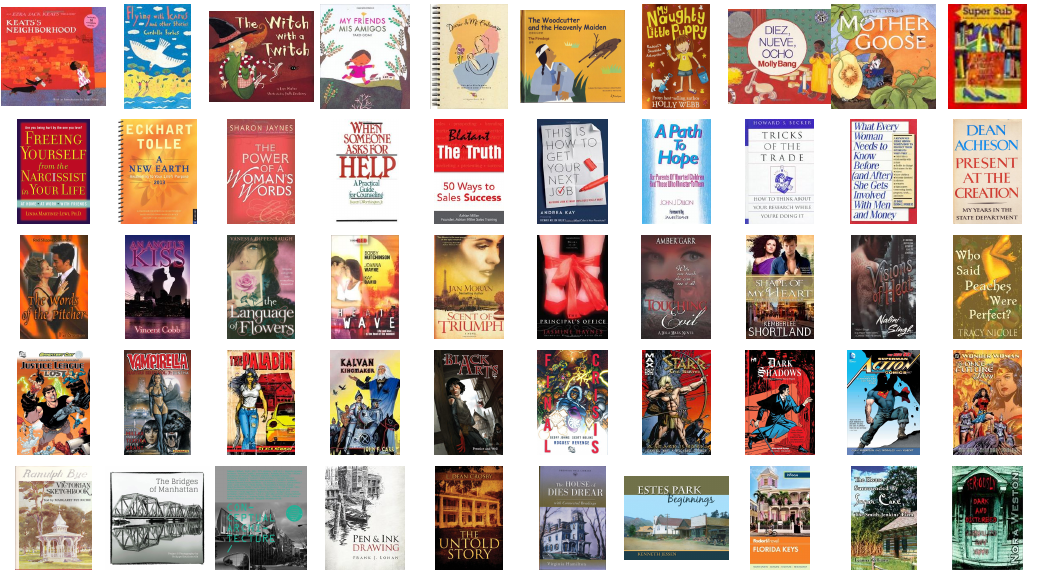}
\end{center}
\normalsize
\vspace{-3mm}
\caption{Examples of K-means clusters in style space (Books `also viewed' and `also bought' data). Although `styles' for categories like books are not so readily interpretable as they are for clothes, visual features are nevertheless able to uncover meaningful distinctions between different product categories, e.g.~the first four rows above above appear to be children's books, self-help books, romance novels, and graphic novels.\label{fig:pca_books}}
\end{figure}

For each category, we consider the subset of relationships from $\E$ that connect products within that category. After generating random samples of non-relationships, we separate $\E$ and $\NegSet$ into training, validation, and test sets (80/10/10\%, up to a maximum of two million training relationships). Although we do not fit hyperparameters (and therefore do not make use of the validation set), we maintain this split in case it proves useful to those wishing to benchmark their algorithms on this data. While we did experiment with simple $\ell_2$ regularizers, we found ourselves blessed with a sufficient overabundance of data that overfitting never presented an issue (i.e., the validation error was rarely significantly higher than the training error).

To be completely clear, our protocol consists of the following:
\begin{enumerate}
 \item Each category and graph type forms a single experiment (e.g.~predict `bought together' relationships for Women's clothing).
 \item Our goal is to distinguish relationships from non-relati\-onships (i.e., link prediction). Relationships are identified when our predictor \eq{eq:prob} outputs $P(r_{ij} \in \E) > 0.5$.
 \item We consider \emph{all} positive relationships and a random sample of non-relationships (i.e., `distractors') of equal size. Thus the performance of a random classifier is 50\% for all experiments.
 \item All results are reported on the test set.
\end{enumerate}

Results on a selection of top-level categories are shown in Table \ref{tab:toplevel}, with further results for clothing data shown in Table \ref{tab:clothing}. Recall when interpreting these results that the learned model has reference to the object images only.  It is thus estimating the existence of a specified form of relationship purely on the basis of appearance.

In every case the proposed method outperforms both the category-based method and weighted nearest neighbor, and the increase from $K=10$ to $K=100$ uniformly improves performance.  Interestingly, the performance on compliments vs.\ substitutes is approximately the same.  The extent to which the $K=100$ results improve upon the WNN results may be seen as an indication of the degree to which  visual similarity between images fails to capture a more complex human visual notion of which objects might be seen as being substitutes or compliments for each other.  This distinction is smallest for `Books' and greatest for `Clothing Shoes and Jewelery' as might be expected.

We have no ground truth relating the true human visual preference for pairs of objects, of course, and thus evaluate above against our dataset.  This has the disadvantage that the dataset contains all of the Amazon recommendations, rather than just those based on decisions made by humans on the basis of object appearance.  This means that in addition to documenting the performance of the proposed method, the results may also be taken to indicate the extent to which visual factors impact upon the decisions of Amazon customers.  The comparison across categories is particularly interesting.  It is to be expected that appearance would be a significant factor in Clothing decisions, but it was not expected that they would be a factor in the purchase of Books.  One possible interpretation of this effect might be that customers have preferences for particular genres of books and that individual genres have characteristic styles of covers.

\subsection{Personalized recommendations}

Finally we evaluate the ability of our model to personalize co-purchasing recommendations to individual users, that is we examine the effect of the user personalization term in (eqs.~\ref{eq:personal1} and \ref{eq:personal2}). Here we do not use the graphs from Tables \ref{tab:toplevel} and \ref{tab:clothing}, since those are `population level' graphs which are not annotated in terms of the individual users who co-purchased and co-browsed each pair of products. Instead for this task we build a dataset of co-purchases from products that users have \emph{reviewed}. That is, we build a dataset of tuples of the form $(i,j,u)$ for pairs of products $i$ and $j$ that were purchased by user $u$. We train on users with at least 20 purchases, and randomly sample 50 co-purchases and 50 non-co-purchases from each user in order to build a balanced dataset. Results are shown in Table \ref{tab:personal}; here we see that the addition of a user personalization term yields a small but significant improvement when predicting co-purchases (similar results on other categories withheld for brevity).

\begin{figure}[t]
\setlength{\tabcolsep}{0.5mm}
\includegraphics[width=\linewidth]{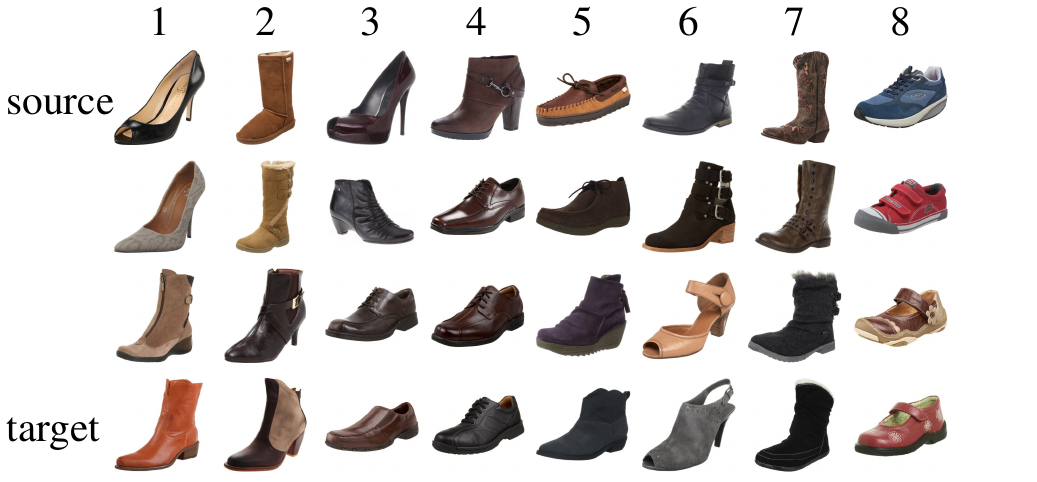}
\caption{Navigating to distant products: each \emph{column} shows a low-cost path between two objects such that adjacent products in the path are visually consistent, even when the end points are not. \label{fig:nav}}
\end{figure}

\begin{figure}[t]
	\centering
\framebox{\adjustbox{trim={.22\width} {.15\height} {.1\width} {.1\height},clip}{\includegraphics[width=1.4\linewidth]{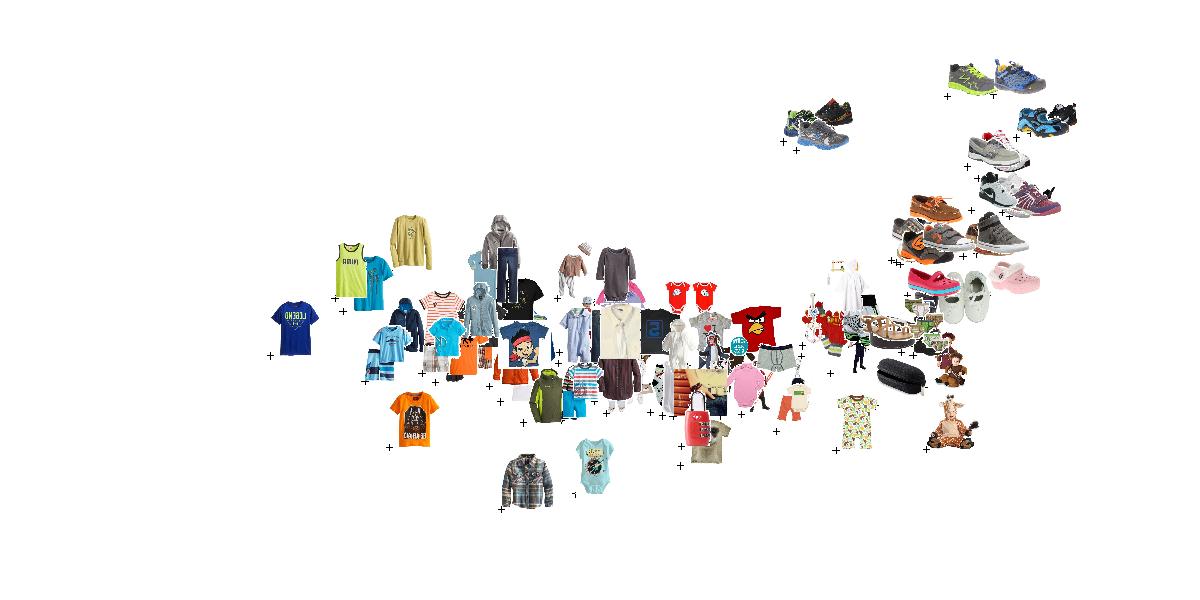}}}
	\caption{A 2-dimensional embedding of a small sample of Boys clothing images (`also viewed' data).}
	\label{fig:stylegraph}
\end{figure}

\section{Visualizing Style Space}

Recall that each image is projected into `style-space' by the transformation $\s_i = \x_i \U$, 
and note that 
the fact that it is based on pairwise distances alone means that the embedding is invariant under isomorphism. That is, applying rotations, translations, or reflections to $\s_i$ and $\s_j$ will preserve their distance in \eq{eq:styledist}. 
In light of these factors we perform k-means clustering on the $K$ dimensional embedded coordinates of the data in order to visualize the effect of the embedding.

Figure \ref{fig:pca} shows images whose projections are close to the centers of a set of selected representative clusters for Men's and Women's clothing (using a model trained on the `also viewed' graph with $K=100$). Naturally items cluster around colors and shapes (e.g. shoes, t-shirts, tank tops, watches, jewelery), but more subtle characterizations exist as well. For instance, leather boots are separated from ugg (that is sheep skin) boots, despite the fact that the visual differences are subtle. This is presumably because these items are preferred by different sets of Amazon users. Watches cluster into different color profiles, face shapes, and digital versus analogue. Other clusters cross multiple categories, for instance we find clusters of highly-colorful items, items containing love hearts, and items containing animals. Figure~\ref{fig:WhiteShoe} shows a set of images which project to locations that span a cluster.

Although performance is admittedly not outstanding for a category such as books, it is somewhat surprising that an accuracy of even 70\% can be achieved when predicting book co-purchases. Figure \ref{fig:pca_books} visualizes a few examples of style-space clusters derived from Books data. Here it seems that there is at least some meaningful information in the cover of a book to predict which products might be purchased together---children's books, self-help books, romance novels, and comics (for example) all seem to have characteristic visual features which are identified by our model.

In Figure \ref{fig:nav} we show how our model can be used to navigate between related items---here we randomly select two items that are unlikely to be co-browsed, and 
find a low cost path between them as measured by our learned distance measure.
Subjectively, the model identifies visually smooth transitions between the source and the target items.

Figure~\ref{fig:stylegraph} provides a visualization of the embedding of Boys clothing achieved by setting $K=2$ (on co-browsing data). Sporting shoes drift smoothly toward slippers and sandals, and underwear drifts gradually toward shirts and coats.

\begin{figure}[t]
\begin{center}
\setlength{\tabcolsep}{0.5mm}
\small
\includegraphics[width=\linewidth]{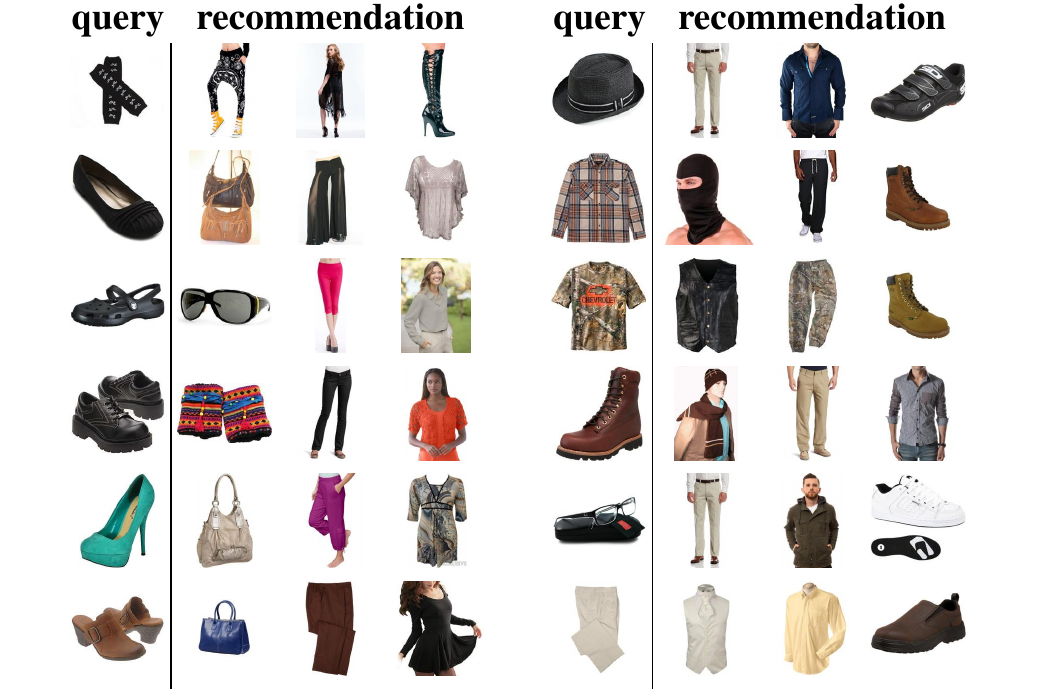}
\normalsize
\setlength{\tabcolsep}{6pt}
\end{center}
\caption{Outfits generated by our algorithm (Women's outfits at left; Men's outfits at right). The first column shows a `query' item that is randomly selected from the product catalogue. The right three columns match the query item with a top, pants, shoes, and an accessory, 
(minus whichever category contains the query item).
\label{fig:outfits}}
\end{figure}

\begin{figure*}[ht]
\renewcommand{\tabcolsep}{0.25mm}
\begin{center}
\includegraphics[width=\linewidth]{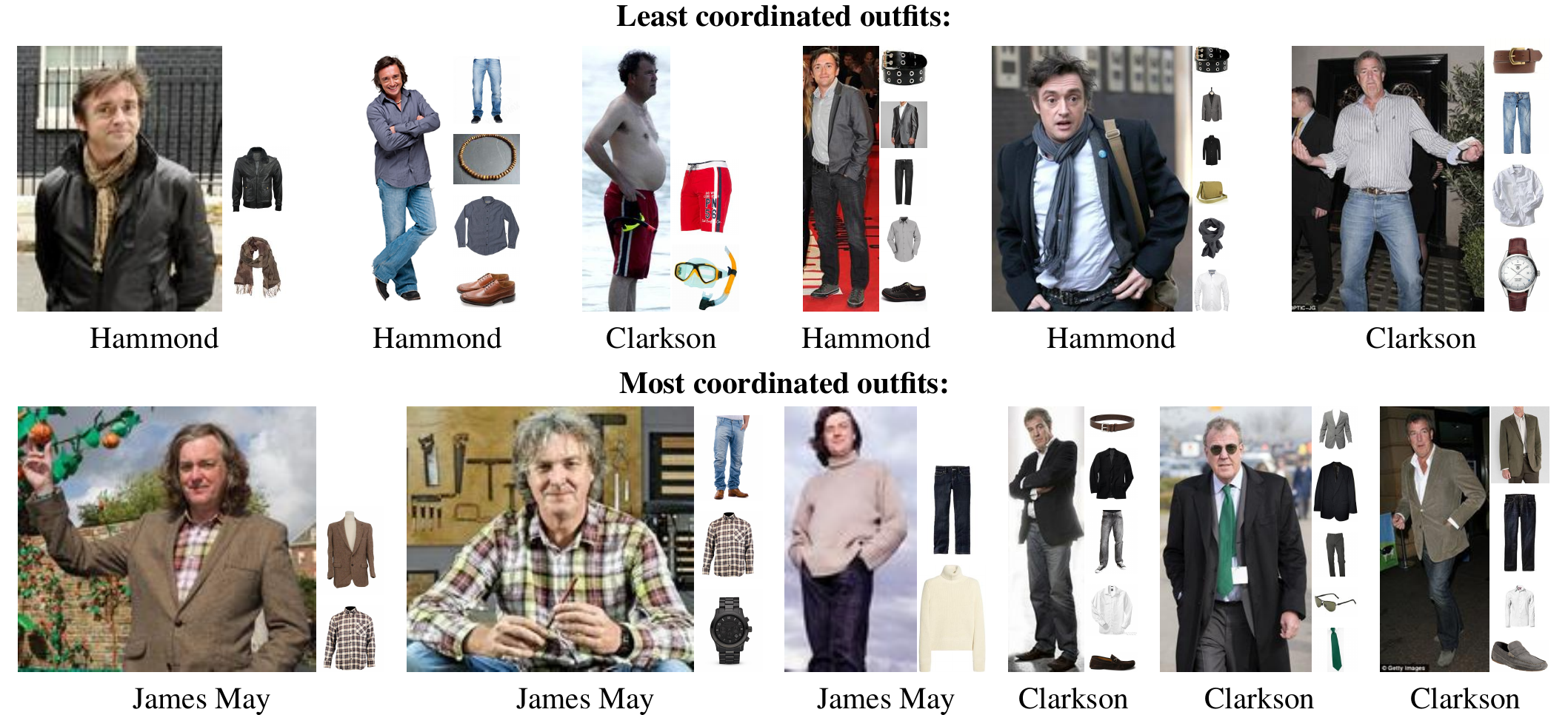}
\end{center}
\caption{Least (top) and most (bottom) coordinated outfits from our Top Gear dataset. Richard Hammond's outfits typically have low coordination, James May's have high coordination, and Jeremy Clarkson straddles both ends of the coordination spectrum. Pairwise distances are normalized by the number of components in the outfit so that there is no bias towards outfits with fewer/more components.\label{fig:topgear}}
\end{figure*}

\begin{figure*}[ht]
\renewcommand{\tabcolsep}{0.47mm}
\begin{center}
\small
\includegraphics[width=\linewidth]{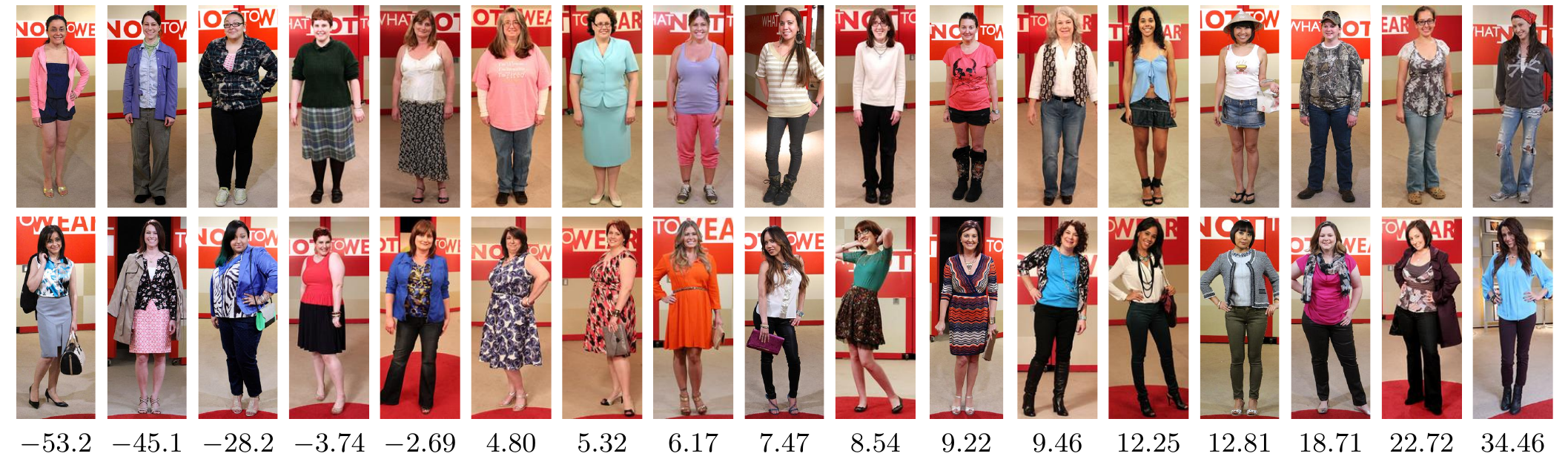}
\normalsize

\end{center}

\renewcommand{\tabcolsep}{6pt}
\vspace{-1mm}
\caption{Contestants in \emph{What Not to Wear}. Original outfits (top), `made-over' outfits (bottom), and the change in log-likelihood ($\delta$) between the components of the old and the new outfits (positive $\delta$ denotes an increase in coordination). \label{fig:wnw}}
\end{figure*}

\section{Generating Recommendations}

We here demonstrate that the proposed model can be used to generate recommendations that might be useful to a user of a web store. Given a query item (e.g.~a product a user is currently browsing, or has just purchased), our goal is to recommend a selection of other items that might complement it. For example, if a user is browsing pants, we might want to recommend a shirt, shoes, or accessories that belong to the same style.

Here, Amazon's rich and detailed category hierarchy can help us. For categories such as women's or men's clothing, we might define an `outfit' as a combination of pants, a top, shoes, and an accessory (we do this for the sake of demonstration, though far more complex combinations are possible---our category tree for clothing alone has hundreds of nodes). Then, given a query item our goal is simply to select items from each of these categories that are most likely to be connected based on their visual style.

Specifically, given a query item $\x_q$, for each category $\C$ (represented as a set of item indices), we generate recommendations according to
\begin{equation}
 \argmax_{j \in \C} \P_\U(\rqj\in \E),
\end{equation}
i.e., the minimum distance according to our measure \eq{eq:U} amongst objects belonging to the desired category.
Examples of such recommendations are shown in Figures~\ref{fig:brownLoaferQuery} and~\ref{fig:outfits}, with randomly chosen queries from women's and men's clothing. Generally speaking the model produces apparently reasonable recommendations, with clothes in each category usually being of a consistent style.

\section{Outfits in The Wild}

An alternate application of the model is to make assessments about outfits (or otherwise combinations of items) that we observe `in the wild'. 
That is, to the extent that the tastes and preferences of \Amazon customers reflect the zeitgeist of society at large, this can be seen as a measurement of whether a candidate outfit is well coordinated visually.  

To assess this possibility, we have built two small datasets of real outfits, one consisting of twenty-five outfits worn by the hosts of \emph{Top Gear} (Jeremy Clarkson, Richard Hammond, and James May), and another consisting of seventeen `before' and `after' pairs of outfits from participants on the television show \emph{What Not to Wear} (US seasons 9 and 10). For each outfit, we cropped each clothing item from the image, and then used \emph{Google}'s reverse image search to identify images of similar items (examples are shown in Figure \ref{fig:topgear}).

Next we rank outfits according to the average log-likeli\-hood of their pairs of components being related using a model trained on Men's/Women's co-purchases (we take the average so that there is no bias toward outfits with more or fewer components). All outfits have at least two items.\footnote{Our measure of coordination is thus undefined for a subject wearing only a single item, though in general such an outfit would be a poor fashion choice in the opinion of the authors.} Figure \ref{fig:topgear} shows the most and least coordinated outfits on \emph{Top Gear}; here we find considerable separation between the level of coordination for each presenter; Richard Hammond is typically the least coordinated, James May the most, while Jeremy Clarkson wears a combination of highly coordinated and highly uncoordinated outfits.

A slightly more quantitative evaluation comes from the television show \emph{What Not to Wear}: here participants receive an `outfit makeover', hopefully meaning that their made-over outfit is more coordinated than the original. Examples of participants before and after their makeover, along with the change in log likelihood are shown in Figure \ref{fig:wnw}. Indeed we find that made-over outfits have a higher log likelihood in 12 of the 17 cases we observed ($p \simeq 7\%$; log-likelihoods are normalized to correct any potential bias due to the number of components in the outfit). This is an important result, as it provides external (albeit small) validation of the learned model which is independent of our dataset.

\section{Conclusion}

We have shown that it is possible to model the human notion of what is visually related by investigation of a suitably large dataset, even where that information is somewhat tangentially contained therein.
We have also demonstrated that the proposed method is capable of modeling a variety of visual relationships beyond simple visual similarity.  
Perhaps what distinguishes our method most is thus its ability to model what makes items \emph{complementary}. To our knowledge this is the first attempt to model human preference for the appearance of one object given that of another in terms of more than just the visual similarity between the two.  It is almost certainly the first time that it has been attempted directly and at this scale.

We also proposed visual and relational recommender systems as a potential problem of interest to the information retrieval community, and provided a large dataset for their training and evaluation.
In the process we managed to figure out what not to wear, how to judge a book by its cover, and to show that James May is more fashionable than Richard Hammond.
\small
\ \\
\textbf{Acknowledgements.} This research was supported by the Data 2 Decisions Cooperative Research Centre, and the Australian Research Council Discovery Projects funding scheme DP140102270.
\normalsize

\footnotesize
\bibliographystyle{abbrv}

\begin{flushleft}

\end{flushleft}

\end{document}